\documentclass[11pt]{article}
\usepackage{coling2020}
\usepackage{times}
\usepackage{url}
\usepackage{hyperref}
\hypersetup{
    colorlinks=false,
    linkcolor=blue,
    filecolor=magenta,      
    urlcolor=cyan,
}
\usepackage{latexsym}
\usepackage{amsmath}
\usepackage{todonotes}
\usepackage[utf8]{inputenc}
\usepackage{multirow}
\usepackage{arydshln}
\usepackage{subcaption}
\colingfinalcopy %

\title{Mining Large-Scale Low-Resource Pronunciation Data From Wikipedia}
\author{Tania Chakraborty, Manasa Prasad, Theresa Breiner, Sandy Ritchie, Daan van Esch \\
Google Research\\
  {\tt \{taniarini, pbmanasa, tbreiner, sandyritchie, dvanesch\}@google.com}}
\date{June 2020}

\begin{document}

\maketitle

\begin{abstract}
Pronunciation modeling is a key task for building speech technology in new languages, and while solid grapheme-to-phoneme (G2P) mapping systems exist, language coverage can stand to be improved. The information needed to build G2P models for many more languages can easily be found on Wikipedia, but unfortunately, it is stored in disparate formats. We report on a system we built to mine a pronunciation data set in 819 languages from loosely structured tables within Wikipedia. The data includes phoneme inventories, and for 63 low-resource languages, also includes the grapheme-to-phoneme (G2P) mapping. 54 of these languages do not have easily findable G2P mappings online otherwise. We turned the information from Wikipedia into a structured, machine-readable TSV format, and make the resulting data set publicly available so it can be improved further and used in a variety of applications involving low-resource languages.
\end{abstract}

\section{Introduction}
\label{intro}

There are thousands of languages spoken around the world, and many efforts to learn about them and document them. However, information about low-resource languages is not always easy to find, or leveraged to its full potential. Wikipedia is a useful resource for language-specific information, with a community of native speakers and linguistic experts who continue to improve coverage and quality\footnote{See the LingWiki effort: \url{https://en.wikipedia.org/wiki/Wikipedia:GLAM/SOAS/Lingwiki}}. While Wikipedia is often used to build monolingual or parallel text corpora \cite{prasad-2018,rahma-2018}, its articles on specific languages contain useful information that is not frequently extracted.

In this paper we describe how we extracted pronunciation information from these Wikipedia pages for hundreds of languages, some of which are not represented in other sources, which can be used in technologies for low-resource languages. The data was mined from tables containing information such as phonemes, associated graphemes, and sample words in each language. While the tables are understandable to the human eye, it is nontrivial to automatically parse and generate standardized TSV files from them. We publish our mined data and encourage others to leverage and improve this source.

\section{Uses for Pronunciation Data}
\label{uses}
While large-scale pronunciation data can be useful in many linguistic pursuits, grapheme-to-phoneme (G2P) mappings are also a building block for automatic language processing systems. Tools to generate pronunciations from input words \cite{epitran-2018,novak-2016} typically support a specific set of languages but can often be extended using G2P rules for new languages. G2P data can also be used in cross-lingual transfer learning, e.g. in named entity recognition \cite{bharadwaj-etal-2016}.

G2P mappings are also required in typical automatic speech recognition (ASR) and text-to-speech (TTS) systems. While these systems usually train on paired audio and transcription data, several recent areas of research leverage G2P mappings to overcome this requirement. If there is a G2P mapping and some text available in a low-resource language, an ASR system can be built by repurposing an acoustic model from a similar higher-resource language \cite{prasad-2019}. G2P mappings can even be bootstrapped for new languages using only the language's phoneme inventory combined with higher-resource language text written in the same script \cite{bleyan-etal-2019}, making even simple phoneme data a more useful tool for scaling technologies to new languages.

\section{Similar Resources}
\label{similar}
One similar resource in this space is PHOIBLE \cite{Phoible14}, a database listing the phonemes of over 2,100 languages, their possible allophones, and how common each phoneme is in the language, but not any G2P information. Wikimedia's Wiktionary and Incubator\footnote{\url{https://meta.wikimedia.org/wiki/Wiktionary\#List_of_Wiktionaries}} are open-content dictionaries written in about 500 languages, and can be used to automatically extract phoneme and pronunciation information \cite{schlippe-2010,lee-etal-2020}. Deri and Knight \shortcite{deri-2016} as well as Peters et al. \shortcite{peters-2017} have been able to adapt this data even to build G2P models for new languages that do not appear in the mined data.

Deri and Knight also published the G2P mappings that they mined from Wikipedia IPA Help tables\footnote{\url{https://en.wikipedia.org/wiki/Category:International_Phonetic_Alphabet_help}} for 98 languages. Our data set, which includes phoneme inventories for over 800 languages, offers G2P data in 63 low-resource languages, 54 of which are not covered by Deri and Knight. There is also more potential G2P information stored for an additional 224 languages, as we detail in section \ref{challenges}. We hope that this additional data set will be a useful supplement to the existing resources.

\section{Mining Pronunciation Data from Wikipedia}
\label{mining}

The English Wikipedia contains hundreds of articles about specific languages\footnote{\url{https://en.wikipedia.org/wiki/Index_of_language_articles}}, often including details on the phonology and writing systems, and covers languages that are not represented in other data sources. We wanted to aggregate this data into an easily comparable and processable data set to make it more widely usable for the research that we described in section \ref{uses}.

\subsection{Extracting the Wikipedia Pages}

Wikipedia's category structure enabled us to target relevant pages inter-linked from the category pages for Languages by Country\footnote{\url{http://en.wikipedia.org/wiki/Category:Languages_by_country}} and Language Phonologies\footnote{\url{http://en.wikipedia.org/wiki/Category:Language_phonologies}}. We also checked for other potential language pages by leveraging the Wikipedia search query for ISO639 codes\footnote{\url{https://iso639-3.sil.org/}}; for example, the page on Amarasi (ISO 639 code `aaz') can be found at \url{https://en.wikipedia.org/wiki/ISO_639:aaz}. Using tooling similar to wikitable2csv\footnote{\url{https://github.com/gambolputty/wikitable2csv}}, we extracted the table data embedded in each targeted page.

\subsection{Parsing the Tables}

From the tables on each language's page, we wanted to parse: a) the \textbf{phoneme inventory} of the language b) the \textbf{pronunciation features} of the phonemes, including voicing, place/manner of articulation, and other phonetic features, useful for clarity and in case of nonstandard transcriptions c) \textbf{grapheme to phoneme} mappings d) \textbf{words} in the native script and e) the phonemic \textbf{transcriptions} of the words.

While some or all of this data is present in most of the language pages, different pages format their tables in different ways, which makes data extraction and processing tricky. We aim to cover as many edge cases as possible and successfully parse a majority of the table formats. Some especially odd cases and ones that we left unaddressed can be found in section \ref{challenges}.

\begin{figure}[h!]
\begin{subfigure}[t]{0.5\textwidth}
\centering
\includegraphics[scale=0.35]{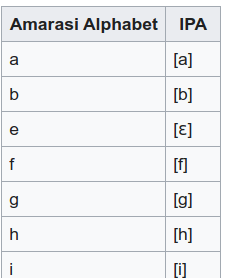}
\caption{Amarasi (iso:aaz) grapheme-phoneme table}
\label{fig:amarasi-table}
\end{subfigure}
\begin{subfigure}[t]{0.5\textwidth}
\centering
\includegraphics[scale=0.45]{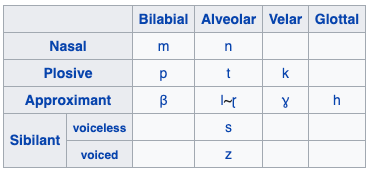}
\caption{Alekano (iso:gah) consonant phoneme table}
\label{fig:mad-table}
\end{subfigure}
\caption{Example Wikipedia tables with relatively simple formats.}
\end{figure}

We mainly saw two different ways that data in a table are presented, which we will call Type A and Type B. \textbf{Type A} has one set of headers, where the contents are related to each other horizontally (or vertically if headers are on the side). For example, Figure \ref{fig:amarasi-table}\footnote{Screenshots of pages accessed on June 24, 2020: \url{https://en.wikipedia.org/wiki/Amarasi_language} and \url{https://en.wikipedia.org/wiki/Alekano_language}} is a Type A table where the data in the column titled ``Amarasi Alphabet'' are graphemes, and the data in the column titled IPA are the corresponding phonemes. \textbf{Type B} has two sets of headers, in which case the data between columns is not related, but rather classified in a similar way. For example, in Figure \ref{fig:mad-table}, the phonemes in the columns are not related to each other; the data in each cell is classified by two or more pronunciation features, namely place of articulation (column headers) and manner of articulation/voicing (row headers).

Since there is no obvious indicator of a table's type from its representation, we have to automatically determine how to parse the data by observing the structure of the table; i.e., how many headers the table has, if the table has both column and row headers, etc. We determine that a given row or column is a header if it has consecutive `header' type cells, which are annotated with a `!' symbol\footnote{According to the Wikipedia Table Help page: \url{https://en.wikipedia.org/wiki/Help:Table}}. We treat tables with headers only on the top or the side as Type A, while tables with both are Type B.

\begin{table}[]
\begin{center}
\begin{tabular}{|l|l|}
\hline
Category & Targeted Keywords \\
\hline
Grapheme & letter, grapheme, alphabet, written \\
Phoneme & IPA, pronunciation \\
Pronunciation Feature & description, vowel/consonant [found in table caption rather than header] \\
Example word & example, word \\
Transcription & transcription \\
Unclassified & [any unmatchable data] \\
\hline
\end{tabular}
\caption{Keywords in table headers that helped us determine the category of data in the table.}
\label{tab:keywords}
\end{center}
\end{table}

The data extracted from the tables is classified into one of the categories listed in Table \ref{tab:keywords}. The correct category for each data item is based on the type of the table and the text in the header corresponding to the item's column. For Type A tables, a regular expression is used to determine the category, based on keywords as you can see in Table \ref{tab:keywords}. If the header text contains any variation of the keywords we are searching for, we can classify that data into the matching category.

For Type B tables, the data is classified as a pronunciation feature, with the cell value representing the associated phoneme and the actual pronunciation features extracted from the row and column headers.

\subsection{Challenges}
\label{challenges}

\begin{figure}[h!]
\begin{center}
\includegraphics[scale=0.35]{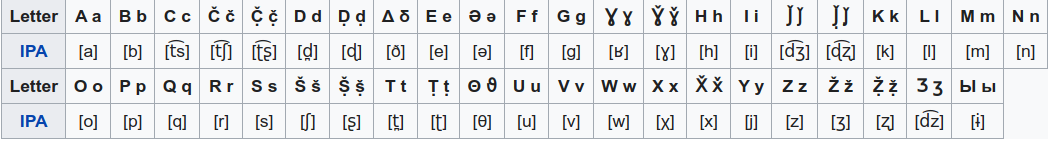}
\caption{Wakhi (iso:wbl) data table with repeated headers across rows}
\label{fig:repeated-rows}
\end{center}
\end{figure}

As we described in Section \ref{mining}, the primary challenge in mining the data was the variety of the table structures. We tried to account for the most common types of tables, including the simpler Type A and Type B tables described above, as well as tables with repeated headers, as seen in Figure \ref{fig:repeated-rows}\footnote{Screenshot of page accessed on June 24, 2020: \url{https://en.wikipedia.org/wiki/Wakhi_language}}. If repetition is not accounted for, the contents of an entire column would be incorrectly associated with each other. Instead, we need to associate only two adjacent rows with each other as we parse the table. The same principle applies to tables that have a repetitions every N rows or every N columns for any value of N. Some more table examples can be found in Appendix \ref{appendix-handle}.

Another challenge is that the keywords in Table \ref{tab:keywords} may be insufficient to classify data correctly for all cases. For example, if a table uses the name of the language as the header for the graphemes in a G2P table instead of a more generic keyword, we will not parse the data correctly. As we developed the parser, we added in new keywords to cover more cases, but do not cover all edge cases. When we find a table whose header is not a keyword, we store the data item and its associated header together in the Unclassified category rather than discarding it. Some examples can be found in Appendix \ref{appendix-handle}.

If a keyword is used in a way we do not expect, or if there is a very intricate table format, there can be other problems. Some tables list multiple data within a cell, and it is not clear what the relationship is in order to split the data or classify it correctly. Some examples of these tables can be found in Appendix \ref{appendix-dont-handle}. Our dataset may include some of this noise, although we hope that in most cases, tables that confused our system would store the data in the Unclassified category to at least be human readable.

\section{Our Data Set}
\label{format}

We found phoneme data for 819 languages, and G2P mappings for 63 languages. 54 of these do not appear in Deri and Knight's G2P data for 98 languages mined from the Wikipedia IPA Help Tables \footnote{\url{https://drive.google.com/drive/u/1/folders/0B7R_gATfZJ2aWkpSWHpXUklWUmM}} \cite{deri-2016}. An additional 224 languages had some data we could only parse as Unclassified, but further analysis hints that 72 of them may contain G2P information. Since the Wikipedia pages for higher-resource languages do not often contain G2P tables (but tend to focus on grammar or history), our data set does not include G2P for these languages, which are more likely covered by Deri and Knight, and instead represents more low-resource languages.

We publish our data \footnote{\url{ https://github.com/google/language-resources/tree/master/mined-wiki-phoneme-tables}} in TSV format, where the columns correspond to: grapheme, phoneme, pronunciation features, example word, IPA transcription of the example word, and Unclassified data. Phonemes can appear in more than one row if, for example, they are mapped to more than one grapheme or example word. If a field is empty, it is filled in with `(n/a)'. See Figure \ref{fig:amarasi-tsv} for an example excerpt, with more examples available in the appendix. We also include a txt with all Wikipedia links that were used.

\begin{figure}[h!]
\begin{center}
\includegraphics[scale=0.5]{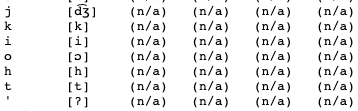}
\caption{Excerpt of Amarasi (iso:aaz) TSV data file, containing g2p mappings}
\label{fig:amarasi-tsv}
\end{center}
\end{figure}

\section{Conclusion}

We mined a data set covering 819 languages' phonemes, including 63 languages' grapheme-to-phoneme mappings, 54 of which are low-resource languages that do not have easily findable G2P mappings in other available sources. This information is available on Wikipedia, but inconsistencies with how it is formatted make it quite challenging to use in extending language technology. While our extraction system might be improved to handle more edge cases in future work, it may in fact be even more fruitful to collaborate with relevant Wikipedia editors to standardize their contributions to make the data more machine-readable. At any rate, we hope this work helps the research community continue to scale research and technologies to more and more languages.

\appendix
\clearpage

\section{Additional Examples of Table Formats We Handle}
\label{appendix-handle}

This appendix gives some further examples of table formats that are nontrivial to parse in order to associate and classify the data correctly.

\begin{figure}[h!]
\begin{center}
\includegraphics[scale=0.35]{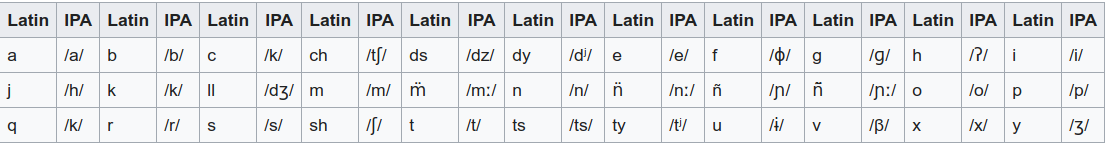}
\caption{Ocaina (iso:oca) data table with repeated headers across columns}
\label{fig:repeated-cols}
\end{center}
\end{figure}

In Figure \ref{fig:repeated-cols} \footnote{Screenshot of page accessed on June 24, 2020: \url{https://en.wikipedia.org/wiki/Ocaina_language}}, we see that tables may have not only repeated row headers (as seen in Figure \ref{fig:repeated-rows}) but may have repeated column headers. We handle this case.

\begin{figure}[h!]
\begin{center}
\includegraphics[scale=0.35]{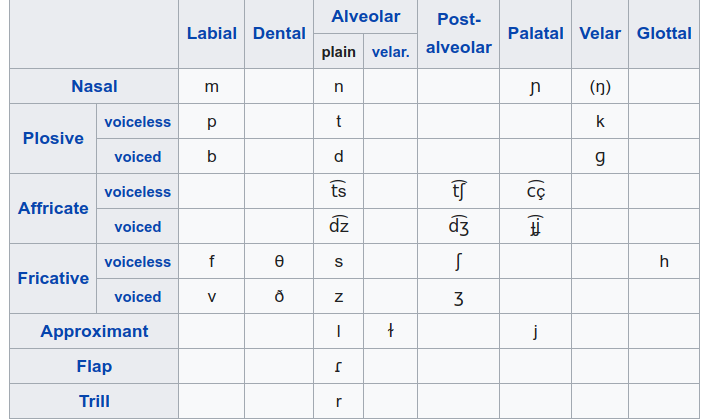}
\caption{Albanian (iso:sq) data table with multiple layers of headers}
\label{fig:multilayered-hdrs}
\end{center}
\end{figure}

In Figure \ref{fig:multilayered-hdrs} \footnote{Screenshot of page accessed on June 24, 2020: \url{https://en.wikipedia.org/wiki/Albanian_language}} every element must be connected with multiple columns and rows of headers to ensure that we do not miss any information. We handle this case.

\begin{figure}[h!]
\begin{subfigure}[t]{0.5\textwidth}
\centering
\includegraphics[scale=0.35]{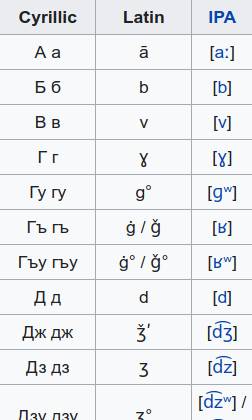}
\caption{Adyghe (iso:ady) data table with multiple orthographies}
\label{fig:multiple-scripts}
\end{subfigure}
\begin{subfigure}[t]{0.5\textwidth}
\centering
\includegraphics[scale=0.35]{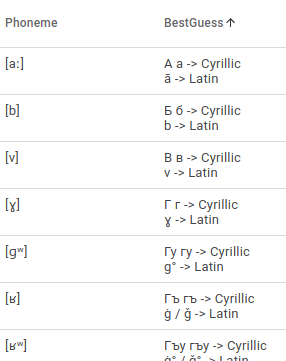}
\caption{Adyghe (iso:ady) extracted data example}
\label{fig:ady_best_guess}
\end{subfigure}
\caption{Adyghe Language extraction example}
\end{figure}

Figure \ref{fig:multiple-scripts} \footnote{Screenshot of page accessed on June 24, 2020: \url{https://en.wikipedia.org/wiki/Adyghe_language}} combines the G2P mappings for two scripts used by the language, with headers that are too specific compared to our keywords for the correct category (grapheme). We handle this case by storing the data and its associated header in the Unclassified category, which is labeled as ``Best Guess" in Figure \ref{fig:ady_best_guess}, and allows the user to see some more information for the language that wasn't straightforward to parse automatically.

\section{Additional Examples of Table Formats We Don't Handle Well}
\label{appendix-dont-handle}

\begin{figure}[h!]
\begin{subfigure}[t]{0.5\textwidth}
\centering
\includegraphics[scale=0.45]{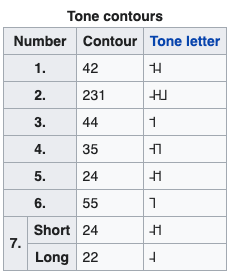}
\caption{E (iso:eee) data table where keyword is \\ used unexpectedly}
\label{fig:keyword-wrong}
\end{subfigure}
\begin{subfigure}[t]{0.5\textwidth}
\centering
\includegraphics[scale=0.45]{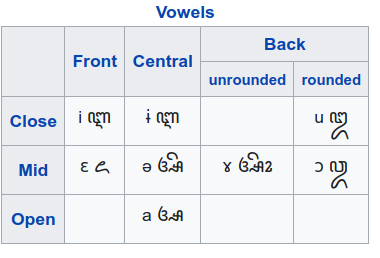}
\caption{Madurese (iso:mad) data table with multiple items per cell}
\label{fig:multiple-data}
\end{subfigure}
\end{figure}

In Figure \ref{fig:keyword-wrong} \footnote{Screenshot of page accessed on June 25, 2020: \url{https://en.wikipedia.org/wiki/E_language}}, a table has a header containing one of the keywords, ``letter", but not in the way that we generally expect, which would be in a table that shows G2P. We do not handle this case and this data is parsed into the data set mistakenly.

In Figure \ref{fig:multiple-data} \footnote{Screenshot of page accessed on June 24, 2020: \url{https://en.wikipedia.org/wiki/Madurese_language}}, there are multiple data items within each cell, and there is no clear way to understand automatically how to split up or classify these items. We do not handle this case.

\section{Additional Example Excerpts from TSV Data Files}

\begin{figure}[h!]
\begin{center}
\includegraphics[scale=0.45]{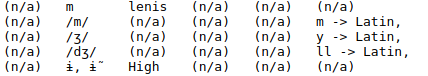}
\caption{Ocaina (iso:oca) tsv format}
\label{fig:oca-tsv}
\end{center}
\end{figure}

\begin{figure}[h!]
\begin{center}
\includegraphics[scale=0.45]{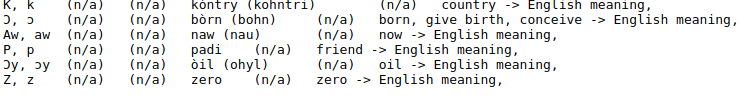}
\caption{Krio (iso:kri) tsv format}
\label{fig:kri-tsv}
\end{center}
\end{figure}

Figure \ref{fig:oca-tsv} is an snippet from the oca.tsv file, and shows how G2P mappings could be captured in the Unclassified column. Figure \ref{fig:kri-tsv} is another example of the type of data that could be captured by the Unclassified column. This language did not have corresponding phonemes in the table, instead it had words from the language and their English meaning.

\clearpage


\begin{thebibliography}{}

\bibitem[\protect\citename{Bharadwaj \bgroup et al.\egroup
  }2016]{bharadwaj-etal-2016}
Akash Bharadwaj, David Mortensen, Chris Dyer, and Jaime Carbonell.
\newblock 2016.
\newblock Phonologically aware neural model for named entity recognition in low
  resource transfer settings.
\newblock In {\em Proceedings of the 2016 Conference on Empirical Methods in
  Natural Language Processing}, pages 1462--1472, Austin, Texas, November.
  Association for Computational Linguistics.

\bibitem[\protect\citename{Bleyan \bgroup et al.\egroup
  }2019]{bleyan-etal-2019}
Harry Bleyan, Sandy Ritchie, Jonas~Fromseier Mortensen, and Daan van Esch.
\newblock 2019.
\newblock Developing pronunciation models in new languages faster by exploiting
  common grapheme-to-phoneme correspondences across languages.
\newblock In {\em Proceedings of Interspeech 2019}.

\bibitem[\protect\citename{Deri and Knight}2016]{deri-2016}
Aliya Deri and Kevin Knight.
\newblock 2016.
\newblock Grapheme-to-phoneme models for (almost) any language.
\newblock In {\em Proceedings of the 54th Annual Meeting of the Association for
  Computational Linguistics (Volume 1: Long Papers)}, pages 399--408, Berlin,
  Germany, August. Association for Computational Linguistics.

\bibitem[\protect\citename{Lee \bgroup et al.\egroup }2020]{lee-etal-2020}
Jackson~L. Lee, Lucas~F.E. Ashby, M.~Elizabeth Garza, Yeonju Lee-Sikka, Sean
  Miller, Alan Wong, Arya~D. McCarthy, and Kyle Gorman.
\newblock 2020.
\newblock Massively multilingual pronunciation modeling with {W}iki{P}ron.
\newblock In {\em Proceedings of The 12th Language Resources and Evaluation
  Conference}, pages 4223--4228, Marseille, France, 05. European Language
  Resources Association.

\bibitem[\protect\citename{Moran \bgroup et al.\egroup }2014]{Phoible14}
Steven Moran, Daniel McCloy, and Richard Wright, editors.
\newblock 2014.
\newblock {\em PHOIBLE Online}.
\newblock Max Planck Institute for Evolutionary Anthropology, Leipzig.

\bibitem[\protect\citename{Mortensen \bgroup et al.\egroup }2018]{epitran-2018}
David~R. Mortensen, Siddharth Dalmia, and Patrick Littell.
\newblock 2018.
\newblock {E}pitran: Precision {G}2{P} for many languages.
\newblock In {\em Proceedings of the Eleventh International Conference on
  Language Resources and Evaluation ({LREC} 2018)}, Miyazaki, Japan, May.
  European Language Resources Association (ELRA).

\bibitem[\protect\citename{Novak \bgroup et al.\egroup }2016]{novak-2016}
Josef~Robert Novak, Nobuaki Minematsu, and Keikichi Hirose.
\newblock 2016.
\newblock Phonetisaurus: Exploring grapheme-to-phoneme conversion with joint
  n-gram models in the wfst framework.
\newblock {\em Natural Language Engineering}, 22(6):907–938.

\bibitem[\protect\citename{Peters \bgroup et al.\egroup }2017]{peters-2017}
Ben Peters, Jon Dehdari, and Josef van Genabith.
\newblock 2017.
\newblock Massively multilingual neural grapheme-to-phoneme conversion.
\newblock In {\em Proceedings of the First Workshop on Building Linguistically
  Generalizable {NLP} Systems}, pages 19--26, Copenhagen, Denmark, 09.
  Association for Computational Linguistics.

\bibitem[\protect\citename{Prasad \bgroup et al.\egroup }2018]{prasad-2018}
Manasa Prasad, Theresa Breiner, and Daan van Esch.
\newblock 2018.
\newblock Mining training data for language modeling across the world’s
  languages.
\newblock In {\em Proceedings of the 6th International Workshop on Spoken
  Language Technologies for Under-resourced Languages (SLTU 2018)}.

\bibitem[\protect\citename{Prasad \bgroup et al.\egroup }2019]{prasad-2019}
Manasa Prasad, Daan van Esch, Sandy Ritchie, and Jonas~Fromseier Mortensen.
\newblock 2019.
\newblock Building large-vocabulary asr systems for languages without any audio
  training data.
\newblock In {\em Proceedings of Interspeech 2019}.

\bibitem[\protect\citename{Rahma \bgroup et al.\egroup }2018]{rahma-2018}
Sellami Rahma, Fatiha Sadat, and Lamia Belguith, 2018.
\newblock {\em Building and Exploiting Domain-Specific Comparable Corpora for
  Statistical Machine Translation}, pages 659--676.
\newblock 01.

\bibitem[\protect\citename{Schlippe \bgroup et al.\egroup }2010]{schlippe-2010}
Tim Schlippe, Sebastian Ochs, and Tanja Schultz.
\newblock 2010.
\newblock Wiktionary as a source for automatic pronunciation extraction.
\newblock In {\em Proceedings of Interspeech 2010}.

\end{thebibliography}
\end{document}